%% file: main.tex
\documentclass[english]{IEEEtran}
\usepackage[T1]{fontenc}
\usepackage[latin9]{inputenc}
\usepackage{array}
\usepackage{amsbsy}
\usepackage{amstext}
\usepackage{amssymb}
\usepackage{graphicx}
\PassOptionsToPackage{normalem}{ulem}
\usepackage{ulem}

\makeatletter

\providecommand{\tabularnewline}{\\}

\usepackage{microtype}
\usepackage{flushend}

\makeatother

\usepackage{babel}
\begin{document}
\global\long\def\model{\mathtt{Deepr}}

\title{$\model$: A Convolutional Net for Medical Records}

\author{Phuoc Nguyen, Truyen Tran, Nilmini Wickramasinghe, Svetha Venkatesh}

\maketitle
\global\long\def\xb{\boldsymbol{x}}
\global\long\def\yb{\boldsymbol{y}}
\global\long\def\eb{\boldsymbol{e}}
\global\long\def\zb{\boldsymbol{z}}
\global\long\def\hb{\boldsymbol{h}}
\global\long\def\ab{\boldsymbol{a}}
\global\long\def\bb{\boldsymbol{b}}
\global\long\def\cb{\boldsymbol{c}}
\global\long\def\sigmab{\boldsymbol{\sigma}}
\global\long\def\gammab{\boldsymbol{\gamma}}
\global\long\def\alphab{\boldsymbol{\alpha}}
\global\long\def\rb{\boldsymbol{r}}
\global\long\def\fb{\boldsymbol{f}}
\global\long\def\ib{\boldsymbol{i}}

\begin{abstract}
\input{abs.tex}
\end{abstract}

\section{Introduction}

\input{intro.tex}

\section{Background}

\input{bg.tex}

\section{$\protect\model$: A \uline{Deep} Net for Medical \uline{R}ecords}

\input{method.tex}

\section{Implementation}

\input{implement.tex}

\section{Results}

\input{exp_v2.tex}

\section{Discussion}

\input{discuss.tex}

\bibliographystyle{plain}

\end{document}

%% file: abs.tex
Feature engineering remains a major bottleneck when creating predictive
systems from electronic medical records. At present, an important
missing element is detecting predictive \emph{regular clinical motifs}
from \emph{irregular episodic records}. We present $\model$ (short
for \uline{Deep} \uline{r}ecord), a new \emph{end-to-end} deep
learning system that learns to extract features from medical records
and predicts future risk automatically. $\model$ transforms a record
into a sequence of discrete elements separated by coded time gaps
and hospital transfers. On top of the sequence is a convolutional
neural net that detects and combines predictive local clinical motifs
to stratify the risk. $\model$ permits transparent inspection and
visualization of its inner working. We validate $\model$ on hospital
data to predict unplanned readmission after discharge. $\model$ achieves
superior accuracy compared to traditional techniques, detects meaningful
clinical motifs, and uncovers the underlying structure of the disease
and intervention space.

%% file: intro.tex
A major theme in modern medicine is \emph{prospective healthcare},
which refers to the capability to estimate the future medical risks
for individuals. These risks can include readmission after discharge,
the onset of specific diseases, and worsening from a condition \cite{snyderman2003prospective}.
Such capability would facilitate timely prevention or intervention
for maximum health impact, and provide a major step toward personalized
medicine. An important data resource in aiding this process are electronic
medical records \cite{jensen2012mining}. Electronic medical records
(EMRs) contain a wealth of patient information over time. Central
to EMR-driven risk prediction is \emph{patient representation}, also
known as feature engineering. Representing an EMR amounts to extracting
relevant historical signals to form a feature vector.

However, feature extraction in EMR is challenging \cite{tran2014framework}.
An EMR typically consists of a sequence of time-stamped visit episodes,
each of which has a subset of coded diagnoses, a subset of procedures,
lab tests and textual narratives. The data is \emph{irregular} \emph{at
patient level}. EMR is episodic \textendash{} events are only recorded
when patients visit clinics, and the time gap between two visits is
largely random. Representing irregular timing poses a major challenge.
EMR varies greatly in length \textendash{} young patients usually
have just one visit for an acute condition, but old patients with
chronic conditions may have hundreds of visits. At the same time,
the data is \emph{regular at local episode level}. Diseases tend to
form clusters (comorbidity) \cite{sharabiani2012systematic} and the
disease progression may be dictated by the underlying biological processes
\cite{wang2014unsupervised}. Likewise treatments may follow a certain
protocol or best practice guideline \cite{huang2013latent}, and there
are well-defined disease-treatment interactions \cite{royston2008interactions}.
These regularities can be thought as clinical motifs. Thus an effective
EMR representation should be able to identify\emph{ regular clinical
motifs} out of \emph{irregular data}. 

Existing EMR-driven predictive work often relies on high-dimensional
sparse feature representation, where features are engineered to capture
certain regularities of the data \cite{he2014mining,jensen2012mining}
This feature engineering practice is effort intensive and non-adaptive
to varying medical records systems. Automated feature representation
based on bag-of-words (BoW) is scalable, but it breaks collocation
relations between words and ignores the temporal nature of the EMR,
thus it fails to properly address the aforementioned challenges. 

In this work we present a new prediction framework called $\boldsymbol{\model}$
that does not require manual feature engineering. The technology is
based on \emph{deep learning,} a new revolutionary approach that aims
to build a \emph{multilayered neural learning system} like a brain
\cite{lecun2015deep}. When fed with a large amount of raw data, the
system learns to recognize patterns with little help from domain experts.
Deep learning now powers speech recognition in Google Voice, self-driving
cars at Google and Baidu, question answering system at IBM (Watson),
and smart assistants at Facebook. It already has a great impact on
hundreds of millions (if not billions) people. But healthcare has
largely been ignored. We hypothesize that a key to apply deep learning
for healthcare patient representation which requires a proper handling
of the irregular nature of episodes mentioned above \cite{pham2016deepcare}.
$\model$ fills the gap by offering an \emph{end-to-end} technology
that \emph{learns} to represent patients from scratch. It reads medical
records, learns the local patterns, adapts to irregular timing, and
predicts personalized risk.

The architecture of $\model$ is multilayered and is inspired by recent
convolutional neural nets (CNNs) in natural languages \cite{collobert2011natural,kim2014convolutional,lecun2015deep,manning2015computational,zhang2015text}.
The most crucial operation occurs at the bottom level where $\model$
transforms an EMR into a ``sentence'' of multiple phrases separated
by special ``words'' that represent time gap. Each phrase is an
visit episode. As with syntactical grammars and collocation patterns
in NLP, there might exist ``health grammars'' and ``clinical patterns''
in healthcare. Health grammars refer to latent biological and environmental
laws that dictate the global evolution of one's health \emph{over
time,} e.g., probable progression from ``diabetes type II'' to ``renal
failure''. To handle irregular timing, time gaps and transfers are
treated as special words. With this representation, an EMR is transformed
into a sentence of variable length that retains all important events.
The other layers of $\model$ constitute a CNN, which is similar to
those in \cite{collobert2011natural,kim2014convolutional,zhang2015text}.
First, words are embedded into a continuous vector space. Next, words
in sentence are passed through a convolution operation which detects
local motifs. Local motifs are then pooled to form a global feature
vector, which is passed into a classifier, which predicts the future
risk. All components are learned at the same time from data: the data
signals are passed from the data to the output, and the training signals
are propagated back from the labels to the motif detectors. Hence
$\model$ is \emph{end-to-end}.

We validate $\model$ on a large database of 300K patients collected
from a hospital chain in Australia. We focus on predicting \textbf{unplanned
readmission within 6 months} after discharge. Compared to existing
bag-of-words representation, $\model$ demonstrates a superior accuracy
as well as the capacity to learn predictive clinical motifs, and to
uncover the underlying structure of the space of diseases and interventions.

To summarize, we claim the following contributions:
\begin{itemize}
\item A novel representation of irregular-time EMR as a sentence with time
gaps and transfers as special words.
\item A novel deep learning architecture called $\model$ that (i) uncovers
the structure of the disease/treatment space, (ii) discovers clinical
motifs, (iii) predicts future risk and (iv) explains the prediction
by identifying motifs with strong responses in each record. The system
is end-to-end, and its inner working can be inspected and visualized,
allowing interpretability and transparency.
\item An evaluation of these claimed capabilities on a large-scale dataset
of 300K patients.
\end{itemize}

%% file: bg.tex
\paragraph{Medical records}

An electronic medical record (EMR) contains information about patient
demographics and a sequence of hospital visits for a patient. Admission
information may include admission time, discharge time, lab tests,
diagnoses, procedures, medications and clinical narratives. Diagnoses,
procedures and medications are discrete entities. For example, diagnoses
may be represented using ICD-10 coding schemes\footnote{http://apps.who.int/classifications/icd10/browse/2016/en}.
For example, in ICD-10, E10 refers to Type 1 diabetes mellitus, E11
to Type 2 diabetes mellitus. The procedures are typically coded in
CPT (Current Procedural Terminology) or ICHI (International Classification
of Health Interventions) schemes \footnote{http://www.who.int/classifications/ichi/en/}.
One of the most important secondary uses of EMR is building predictive
models \cite{jensen2012mining,mathias2013development,tran2014framework,tran_et_al_kais14}. 

Most existing prediction methods on EMRs either rely on manual feature
engineering \cite{mathias2013development} or simplistic extraction
\cite{tran2014framework}. They either ignore long-term dependencies
or do not adequately capture variable length \cite{arandjelovic2015discovering,mathias2013development,tran2014framework}.
Neither are they able to model temporal irregularity \cite{jackson2003multistate,liu2015temporal,tran2014framework,wang2014unsupervised}.
Capturing disease progression has been of great interest \cite{jensen2014temporal,liu2015temporal},
and much effort has been spent on Markov models \cite{henriques2014generative,jackson2003multistate,wang2014unsupervised}.
As Markov processes are memoryless, Markov models forget severe conditions
of the past when it sees an admission due to common cold. This is
undesirable. A proper modeling, therefore, must be non-Markovian and
able to capture long-term dependencies. 

\paragraph{Deep learning}

Deep learning is an approach in machine learning, aiming at producing
\emph{end-to-end} systems that learn from raw data and perform desired
tasks without manual feature engineering. The current wave of deep
learning was initiated by the seminal work of \cite{hinton2006rdd}
in 2006, but deep learning has been developed for decades \cite{schmidhuber2015deep}.
Over the past few years, deep learning has broken records in cognitive
domains such as vision, speech and natural languages \cite{lecun2015deep}.
Current deep learning is mostly based on multilayered neural networks
\cite{schmidhuber2015deep}. All the networks share a common unit
\textendash{} the neuron \textendash{} which is a simple computational
device that applies a nonlinear transform to a linear function of
inputs: i.e., $f(x)=\sigma\left(b+\sum_{i}w_{i}x_{i}\right)$. Almost
all networks thus far are trained using back-propagation \cite{williams1986learning},
thus enable end-to-end learning.

There are three main deep neural architectures in practice: \emph{feedforward},
\emph{recurrent} and \emph{convolutional}. Feedforward nets (FFN)
pass unstructured information from one end to the other, usually from
an input to an output, hence they act as a universal function approximator
\cite{hornik1989multilayer}. Recurrent nets (RNN) model dynamics
over time (and space) using self-replicated units. They maintain some
degree of memory, and thus have potential to capture long-term dependencies.
RNNs are powerful computational machines \textendash{} they can approximate
any program \cite{lin1996learning}. Convolutional nets (CNN) exploit
the repeated local motifs across time and space, and thus are translation-invariant
\textendash{} the capacity often seen in human visual cortex \cite{lecun1995convolutional}.
Local motifs are small piece of data, usually of pre-defined sizes,
e.g., a batch of pixels, or a n-gram of words. CNN is often equipped
with pooling operations to reduce the resolution and enlarge the motifs.

%% file: method.tex
In this section, we describe our deep neural net named $\model$ (short
for \uline{Deep} net for medical \uline{R}ecord) for representing
Electronic Medical Records (EMR) and predicting the future risk. 

\subsection{$\protect\model$ Overview\label{subsec:-Overview}}

\begin{figure*}
\begin{centering}
\includegraphics[width=0.9\textwidth]{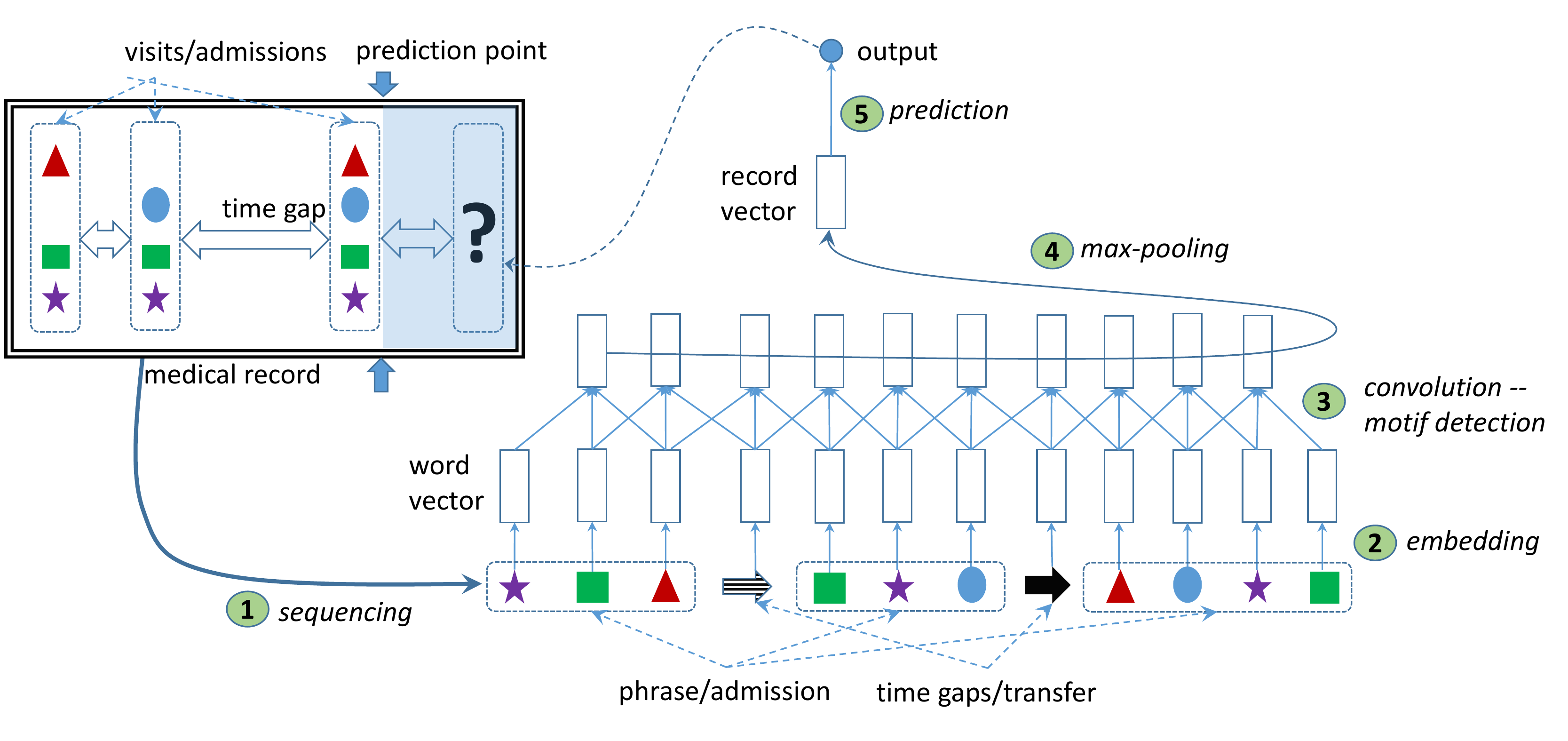}
\par\end{centering}
\caption{Overview of $\protect\model$ for predicting future risk from medical
record. Top-left box depicts an example of medical record with multiple
visits, each of which has multiple coded objects (diagnosis \& procedure).
The future risk is unknown (question mark (?)). \emph{Steps} \emph{from-left-to-right}:
(1) Medical record is sequenced into phrases separated by coded time-gaps/transfers;
then \emph{from-bottom-to-top}: (2) Words are embedded into continuous
vectors, (3) local word vectors are convoluted to detect local motifs,
(4) max-pooling to derive record-level vector, (5) classifier is applied
to predict an output, which is a future event. Best viewed in color.
\label{fig:Overview-of-model}}
\end{figure*}

$\model$ is a multilayered architecture based on convolutional neural
nets (CNNs). The information flow is summarized in Fig.~\ref{fig:Overview-of-model}.
At the bottom level, $\model$\emph{ sequences} the EMR into a ``sentence'',
or equivalently, a sequence of ``words''. Each word represents a
discrete object or event such as diagnosis, procedure, or any derived
object such as time-interval or hospital transfer. The next layer
\emph{embeds} words into an Euclidean space. On top of the embedding
layer is a CNN\emph{ }that reads a small chunk of words in a sliding
window to identify local motifs. The local motifs are transformed
by Rectified Linear Unit (\emph{ReLU}), which is a nonlinear function.
All the transformed motifs are then \emph{max-pooled} across the sentence
to derive an EMR-level feature vector. Finally, a \emph{linear classifier}
is placed at the top layer for prediction. The entire architecture
of $\model$ can be summarized as a function $f(r)$ for record $r$:
\begin{equation}
f(r)\leftarrow\text{Class}\left(\text{Pool}\left\{ \text{ReLU}\left(\text{Conv}\left[\text{Embed}\left\{ \text{Seq}\left(r\right)\right\} \right]\right)\right\} \right)\label{eq:model-function}
\end{equation}

The CNN plays a crucial role as it detects \emph{clinical motifs}
that are predictive. Clinical motifs are co-occurrences of diseases
(also known as comorbidity), disease progression, patterns of disease/treatment,
and patterns of collocating treatments \cite{kim2014convolutional}.
However, as CNN is supervised it requires labels, which may not always
be available (e.g., new patients with short history). A possible enhancement
is through pretraining the embedding layer through a powerful tool
known as \emph{word2vec} \cite{mikolov2013distributed}. As word2vec
is unsupervised and relies on local collocation patterns, clinical
motifs can be pre-detected, and then further refined through CNN with
supervising signals.

\subsection{Sequencing EMR \label{subsec:Sequentializing-EMR}}

This task refers to transforming an EMR into a sentence, which is
essentially a sequence of words. We present here how the words are
defined and arranged in the sentence.

Recall that an EMR is a sequence of time-stamped visit episodes. Each
episode may contain many pieces of information, but for the purpose
of this work, we focus mainly on diagnoses and treatments (which involve
clinical procedures and medications). For simplicity, we do not assume
perfect timing of each piece, and thus an episode is a finite set
of discrete words (diagnoses and treatments). The episode is then
sequenced into a phrase. The order of the element in the phrase may
follow the pre-defined ordering by the EMR system, for example, primary
diagnosis is placed first, followed by secondary diagnoses, followed
by procedures. In absence of this information, we may randomize the
elements.

Within an episode, occasionally, there are one or more transfers between
care providers, for example, separate departments from the same hospital,
or between hospitals. In these cases, an admission is a phrase, and
an episode is a subset of phrases separated by a transfer event. We
create a special word \texttt{\footnotesize{}TRANSFER} for this event.
Between two consecutive episodes, there is a time gap, whose duration
is generally randomly distributed. We discretize the time gap into
five intervals, measured in months: (0-1{]}, (1-3{]}, (3-6{]}, (6-12{]},
12+. Each interval is assigned a unique identifier, which is treated
as a word. For example,\texttt{\small{} 0-1m} is a word for the (0-1{]}
interval gap. With these treatments, an EMR is a sentence of phrases
separated by words for transfers or time gaps. The phrases are ordered
by their natural time-stamps. For robustness, infrequent words are
coded as \texttt{\footnotesize{}RAREWORD}. 

The following is an example of a sentence, where diagnoses are in
ICD-10 format (a character followed by digits), and procedures are
in digits:\\
\\
\begin{tabular}{|>{\raggedright}p{0.95\columnwidth}|}
\hline 
\texttt{\footnotesize{}1910 Z83 911 1008 D12 K31 1-3m R94 RAREWORD
H53 Y83 M62 Y92 E87 T81 RAREWORD RAREWORD 1893 D12 S14 738 1910 1916
Z83 0-1m T91 RAREWORD Y83 Y92 K91 M10 E86 6-12m K31 1008 1910 Z13
Z83.}\tabularnewline
\hline 
\end{tabular}\\

Here the phrases are: {[}\texttt{\footnotesize{}1910 Z83 911 1008
D12}{]}, {[}\texttt{\footnotesize{}R94 RAREWORD H53 Y83 M62 Y92 E87
T81 RAREWORD RAREWORD 1893 D12 S14 738 1910 1916 Z83}{]}, {[}\texttt{\footnotesize{}RAREWORD
Y83 Y92 K91 M10 E86}{]}, and {[}\texttt{\footnotesize{}K31 1008 1910
Z13 Z83}{]}. The time separators are: {[}\texttt{\footnotesize{}1-3m}{]},
{[}\texttt{\footnotesize{}0-1m}{]}, and {[}\texttt{\footnotesize{}6-12m}{]}.
Note that within each phrase, the ordering of words has been randomized.

\subsection{Convolutional Net \label{subsec:Convolutional-Net}}

\paragraph{Embedding}

The first step when applying convolutional nets on a sentence is to
represent discrete words as continuous vectors. One way is to use
the so-called one-hot coding, that is, each word is a binary vector
of all zeros, except for just one position indexed by the word. However,
this representation creates a high-dimensional vector, which may lead
to overfitting and expensive computation. Alternatively, we can use
\emph{word embedding}, which refers to assigning a dense continuous
vector to a discrete word. For example, the second word {[}\texttt{\footnotesize{}Z83}{]}
in the example above may be assigned to 3D vector as (0.1 -2.3 0.5).
In practice, we maintain a look-up table indexed by words, i.e., $E(w)\in\mathbb{R}^{m}$
is the vector for word $w$. The embedding table $E$ is learnable.
Applying word embedding to the sentence yields a sequence of vectors,
where the vector at position $t$ is $\xb_{t}=E(w_{t})$.

\paragraph{Convolution}

On top of the word embedding layers is a \emph{convolutional layer}.
Each convolution operation reads a sliding window of size $2d+1$
and produces $p$ \emph{filter responses} as follows:
\begin{equation}
\zb_{t}=\text{ReLU}\left(\bb+\sum_{j=-d}^{d}W_{j}\xb_{t+j}\right)\label{eq:conv-rect}
\end{equation}
where $\zb_{t}\in\mathbb{R}^{p}$ is filter response vector at position
$t$, $W_{j}\in\mathbb{R}^{p\times m}$ is the convolution kernel
at relative position $j$ (hence, $W\in\mathbb{R}^{p\times m\times(2d+1)}$),
$\bb$ is bias, and $\text{ReLU}(\xb)=\max\left\{ \boldsymbol{0},\xb\right\} $
(element-wise). When it is clear from the context, we use ``filter''
to refer to the learnable device that detects motifs, which are manifestation
of filters in real data. The rectified linear function enhances strong
signals and eliminates weak ones. The bias $\bb$ and the kernel tensor
$W$ are learnable.

\paragraph{Pooling}

Once the local filter responses are computed by the convolutional
layer, we need to \emph{pool} all the responses to derive a global
sentence-level vector. We apply here the max-pooling operator:
\begin{equation}
\bar{\zb}=\max_{t}\left\{ \zb_{t}\right\} \label{eq:pooling}
\end{equation}
where the max is element-wise. Thus the pooled vector $\bar{\zb}$
lives in the same space of $\mathbb{R}^{p}$ as filters responses
$\left\{ \zb_{t}\right\} $. Like the rectifier used in Eq.~(\ref{eq:conv-rect}),
this max-pooling further enhances strong signals across the words
in the sentence.

\paragraph{Classifier}

The final layer of $\model$ is a classifier that takes the pooled
information and predicts the outcome: $f(r)=\text{classifier}(\bar{\zb}(r))$
for record $r$. The main requirement is that the classifiers must
allow gradient to propagate down to lower layers. Examples include
a linear classifier (e.g., logistic regression) or a non-linear parametric
classifier (e.g., neural network).

\subsection{Training}

$\model$ has multiple trainable parameters: embedding matrix, biases,
convolution kernels, and classifier-specific parameters. As the number
of trainable parameters is often large, it necessitates regularizers
such as weight shrinkage (e.g., via $\ell_{2}$ norm) or dropouts
\cite{srivastava2014dropout} . For training we also need to specify
a loss function, which depends on the nature of classifiers. For example,
for binary outcome (e.g., readmission), logistic classifier is usually
trained on cross-entropy loss. Training starts with (random) initialization
of parameters which are then refined through back-propagation and
stochastic gradient descent (SGD). This requires gradients with respect
to trainable parameters. Gradient computation is often tedious and
erroneous, but it is now fully automated in modern deep learning frameworks
such as Theano \cite{bergstra2010theano} and Tensorflow \cite{abadi2016tensorflow}.
For SGD, parameters are updated after every mini-batch of records
(or sentences). Training is stopped after a pre-defined number of
epochs (iterations), or on convergence.

\paragraph{Pretraining with word2vec}

As mentioned in Sec\@.~\ref{subsec:-Overview}, the embedding matrix
can be pretrained using \emph{word2vec. }Here we do not need labels,
and thus we can exploit a large set of unlabeled data.

\subsection{Model Inspection and Visualization \label{subsec:Model-Inspection-and}}

$\model$ facilitates intuitive model inspection and visualization
for better understanding:

\paragraph{Identifying motif responses in a sequence}

For each motif detector, the motifs response at position $t$ (e.g.,
$\zb_{t}\in\mathbb{R}^{p}$) can be used to identify and visualize
strong motifs. For size-3 motifs, the response weight to a size-3
sub-sequence $\left(\xb_{t-1},\xb_{t},\xb_{t+1}\right)$ of a sequence
$\xb$ is the term $\sum_{j=-d}^{d}W_{j}\xb_{t+j}$ in Eq.~(\ref{eq:conv-rect}),
which is the dot product of the sub-sequence and the kernel $W$.

\paragraph{Identifying frequent and strong motifs}

Motifs with large responses in sequences are collected. From this
collection, we keep frequent motifs representative for each outcome
class.

\paragraph{Computing word similarity}

Through embedding $\xb_{w}=E(w)$, word similarity can be computed
easily, e.g., through cosine $S(w,v)=\xb_{w}^{\top}\xb_{v}\left(\left\Vert \xb_{w}\right\Vert \left\Vert \xb_{v}\right\Vert \right)^{-1}$.

\paragraph{Visualization of similar patients}

Patient vectors from Eq.~(\ref{eq:pooling}) can be used to compute
patient similarity. This enables retrieving patients who have similar
history and similar future risk likelihood. This is unlike existing
methods that compute only similar history, which does not necessarily
guarantee similar future. Further, the similarity is not heuristic,
and it does not require a heuristic combination of multiple data types
(such as diseases and interventions). Fig.~\ref{fig:2d-classifications},
for example, shows the distribution of positive and negative classes,
in which patient vectors are projected onto 2D using t-SNE \cite{van2008visualizing}.
Patients who have similar history and future will stay close together. 

\paragraph{Visualization in disease/intervention space}

Since words are embedded into vectors, visualization in 2D is through
dimensionality reduction tools such as PCA or t-SNE \cite{van2008visualizing}.

%% file: implement.tex
In this section, we document implementation details of $\model$ on
a typical EMR system. For ease of exposition, we assume that diseases
are coded in ICD-10 format, but other versions are also applicable
with minimal changes.

\subsection{Data and Evaluation}

Data was collected from a large private hospital chain in Australia
in the period of July 2011 \textendash{} December 2015. The data is
coded according to Australian Coding Standard (ACS). The ACS dictates
that diagnosis coding is based on ICD-10-AM\footnote{https://www.accd.net.au/Icd10.aspx},
an Australian adaptation to WHO's ICD-10 system. Likewise, procedure
coding follows ACHI (Australian Classification of Health Interventions).
The data consists of 590,546 records (300K unique patients), each
corresponds to an admission (defined by an admission time and a discharge
time).

The data subset for testing $\model$ was selected as follows. First
we identified 4,993 patients who had at least an unplanned readmission
within 6 months from a discharge, regardless of the admitting diagnosis.
This constituted the risk group. For each risk case, we then randomly
picked a control case from the remaining patients. For each risk/control
group, we used 830 patients for model tuning, 830 for testing and
the rest for training. A discharge (except for the last one in risk
group) is randomly selected as prediction point, from which the future
risk will be predicted. See also Fig.~\ref{fig:Overview-of-model}
for a graphical illustration.

\subsection{Implementation Details of $\protect\model$}

\paragraph{Episode definition}

$\model$ assumes that episodes are well-defined with an admission
time and discharge time. However, it is not always the case due to
intra-hospital or inter-hospital transfers. Our implementation links
two admissions into an episode if they are separated by less than
12 hours, or by 12-24 hours but with documented transfer.

\paragraph{Words}

For robustness, only level 3 ICD-10-AM codes are used. For example,
\emph{F20.0} (paranoid schizophrenia) would be converted into \emph{F20}
(schizophrenia). Similarly, the procedures are converted into procedure
blocks. Rare words are those occurring less than 100 times in the
database.

\paragraph{Word order randomization}

For motifs detection, randomization is necessary to generate many
potential motifs. We also test a special case where words in a phrase
are ordered starting with the primary diagnosis followed by other
secondary diagnoses, then by procedures in their natural ordering
as defined by the EMR system.

\paragraph{Sentence length}

For CNN, the sentences are trimmed to keep the last min(100, len(sentence))
words. This is to avoid the effects of some patients who have very
long sentences which severely skew the data distribution. In a typical
EMR, this is equivalent to accounting for up-to 10 visits per patient,
which cover more than 95\% of patients.

\paragraph{Hyper-parameter tuning}

$\model$ has a number of hyper-parameters pre-specified by model
users: embedding dimension $m$, kernel window size $2d+1$, motif
size, number of motifs $n$ per size, number of epochs, mini-batch
size, and other classifier-specific settings. Some hyper-parameters
can be found through grid search, which finds the best configuration
with respect to the accuracy on the development set.

We searched for the best parameters using the training and development
data. Then we used the model with the best parameter to predict the
unseen test data. The best parameters settings were $m=100$, $d=1$,
motif size = 3, 4 and 5, $n=100$ number of epochs = 10, mini-batch
size = 64, $\ell_{2}$ regularization $\lambda=1.0$.

\subsection{Baselines}

We implemented the bag-of-words representation and regularized logistic
regression (BoW+LR). LR has a parameter $C$ that helps control overfitting.
We searched for the best parameter $C$ using the development data.
We used the model with the best parameter to predict the unseen test
data. We found the best parameter $C=0.1$, which is equivalent to
a prior Gaussian of mean $0$ and standard deviation of $0.333$.

%% file: exp_v2.tex
\subsection{Risk Prediction}

\begin{table}
\begin{centering}
\begin{tabular}{|l|c|c|}
\hline 
\emph{Method} & \emph{W/o time} & \emph{With time}\tabularnewline
\hline 
BoW + LR & 0.727 & 0.741\tabularnewline
\hline 
\hline 
$\model$ (rand init) & \textbf{0.754} & \textbf{0.753}\tabularnewline
\hline 
$\model$ (\emph{word2vec} init) & \textbf{0.750} & \textbf{0.756}\tabularnewline
\hline 
\end{tabular}
\par\end{centering}
\caption{Accuracy on 6-month unplanned readmission prediction following a random
index discharge with and without time-gaps. Rand init refers to random
initialization of the embedding matrix. \emph{Word2vec} init refers
to pretraining the embedding matrix using the \emph{word2vec} algorithm
\cite{mikolov2013distributed}. \label{tab:unplanned-results} }
\end{table}

We predict unplanned readmission within 6 months after a random index
discharge. Table~\ref{tab:unplanned-results} reports the prediction
accuracy for all methods, when trained on data with and without coded
time-gaps. Time-gaps coding improves the BoW-based prediction, suggesting
the importance of proper sequential handling. However, time-gaps do
not affect the accuracy of $\model$. This might be due to the convolution,
rectification and max-pooling operations (see Sec.~\ref{subsec:Convolutional-Net}),
which pick the most powerful convoluted signals in the sequence. The
use of \emph{word2vec} to initialize the embedding matrix also has
little contribution toward the accuracy. This could be because \emph{word2vec}
looks only for local collocations in both directions (past and future),
whereas the prediction in $\model$ is more global and of longer time
horizon only in the future direction. In either cases with and without
\emph{word2vec}, $\model$ is superior than the baseline BoW+LR. 

\begin{figure*}
\noindent \begin{centering}
\begin{tabular}{cc}
\includegraphics[width=0.95\columnwidth]{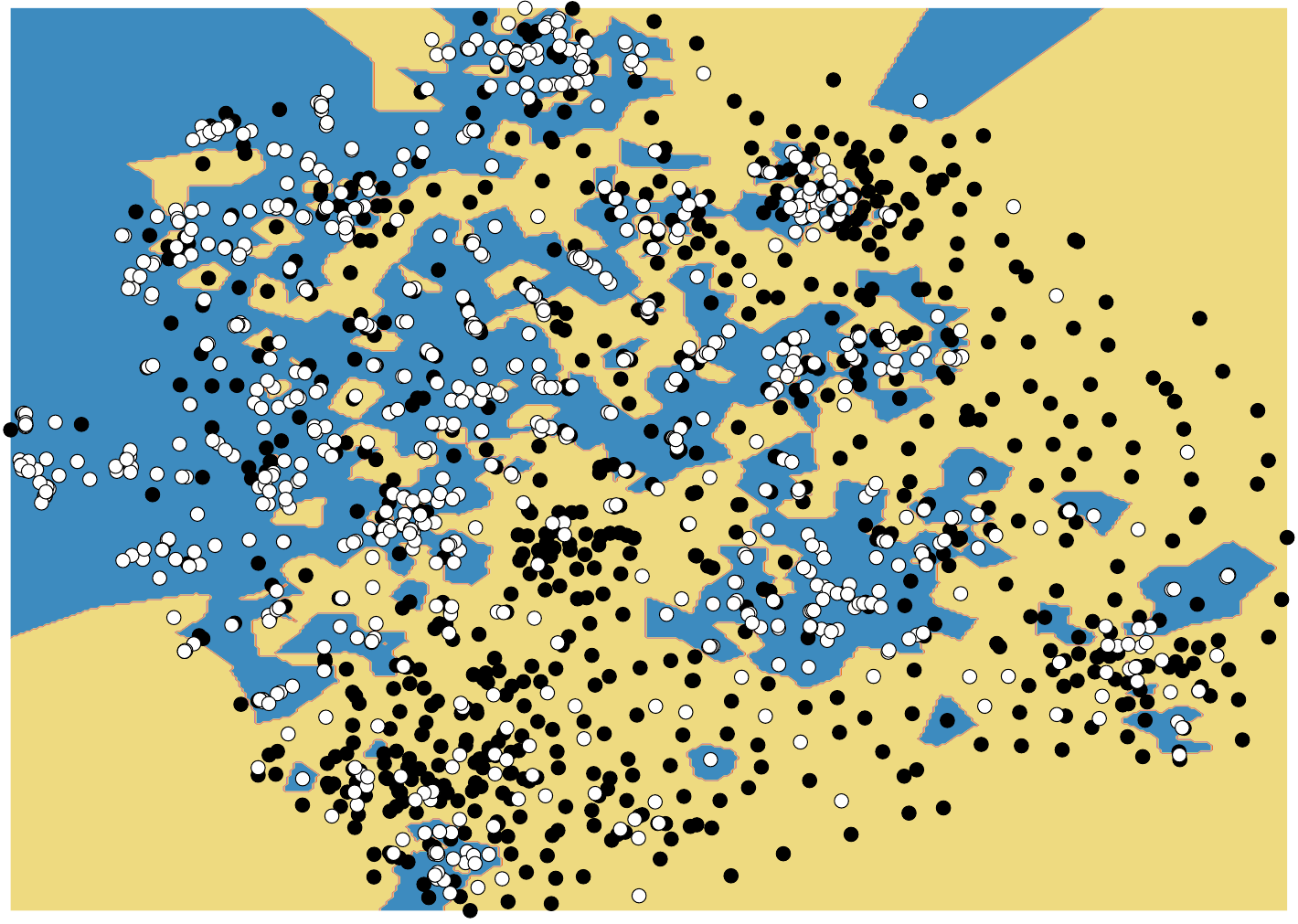} & \includegraphics[width=0.95\columnwidth]{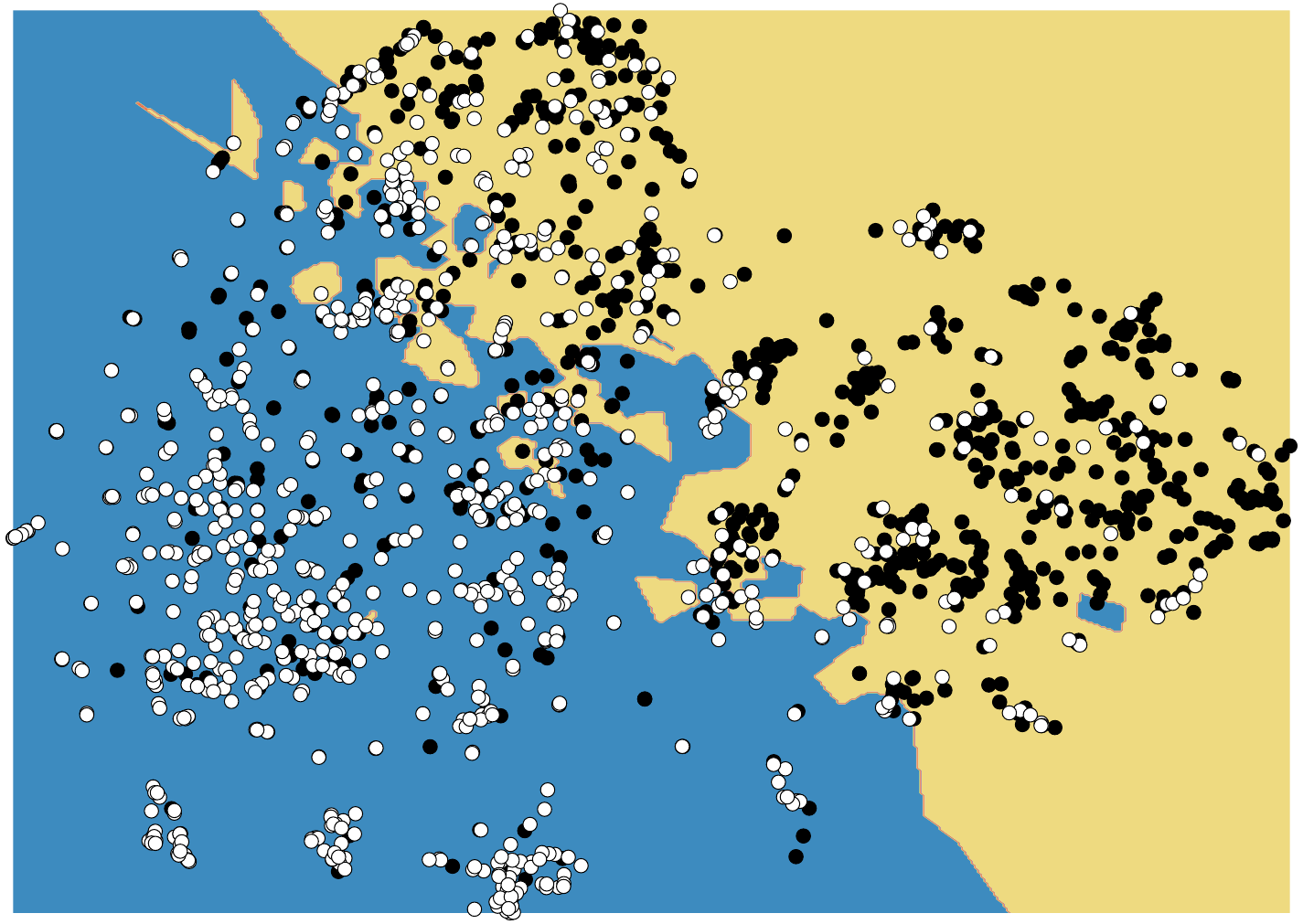}\tabularnewline
BoW+LR & $\model$\tabularnewline
\end{tabular}
\par\end{centering}
\caption{2D projections of classification on the unseen test set of two methods
BoW+LR and $\protect\model$. White points and blue background are
negative class, black point and yellow region are positive class.
The figure shows $\protect\model$ groups similar patients and creates
a more linear decision boundary while BoW+LR scatters the patient
distribution and has a more complicated decision boundary. The decision
boundary is approximated by an exhaustive contouring method, where
fine lattice points of the background grid are labeled to the predicted
label of their nearest data point, and then the boundary is computed
by the contouring algorithm. Best viewed in color.\label{fig:2d-classifications}}

\end{figure*}

Fig.~(\ref{fig:2d-classifications}) shows how $\model$ groups similar
patients and creates a more linear decision boundary while BoW+LR
scatters the patient distribution and has a more complicated decision
boundary. Recall that $\model$ creates the feature vectors using
element-wise max-pooling over all the motifs responses, as in Eq.
~(\ref{eq:pooling}). This demonstrates that the motifs, not just
individual words, are important to computing similarity between patients.
This also suggests that given a new patient $\model$ is better at
querying similar patients in the database when future risk is needed.

\subsection{Disease/Procedure Semantics}

\begin{figure}
\begin{centering}
\includegraphics[width=1\columnwidth]{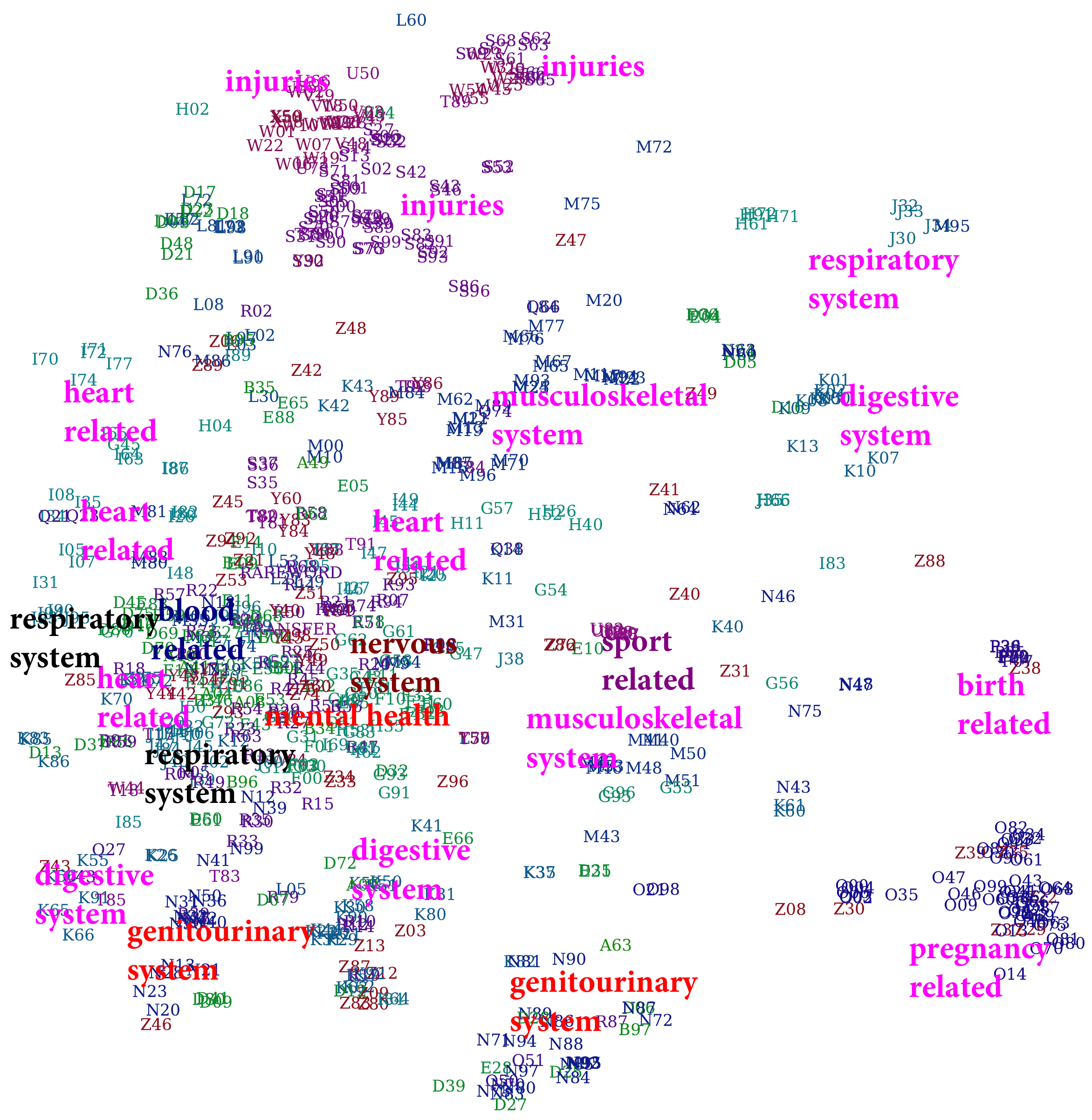}
\par\end{centering}
\caption{Distribution in the disease space, projected into 2D using t-SNE.
Distribution of interventions is omitted for clarity. Best viewed
in color. \label{fig:disease-dist}}

\end{figure}

Recall that $\model$ first embeds words into a vector space. This
offers a simple but powerful way to uncover and visualize the underlying
structure of the word space (see Sec.~\ref{subsec:Model-Inspection-and}).
Fig.~\ref{fig:disease-dist} plots the distribution of diseases on
2D. $\model$ discovers disease clusters which partly correspond to
nodes in the ICD-10 hierarchy. Apart from pregnancy, child birth issues
and injuries, the conditions are not totally separately suggesting
a complex dependencies in the disease space. The main bock of the
disease space has conditions related to heart, blood, metabolic system,
respiratory system, nervous system and mental health. A more close
examination of most similar conditions to a disease is given in Table~\ref{tab:Top-5-randinit}.
For example, similar to cesarean section delivery of baby are those
related to pregnancy complications (disproportion, failed induction
of labor, or diabetes) and corresponding delivery procedures (cesarean
section, manipulating fetal presentation, forceps).

We note in passing that we also obtained a similar visualization using
only \emph{word2vec} as in \cite{mikolov2013distributed}, which is
known to detect hidden semantic relationships between words. $\model$
trained on the embedding matrix initialized by \emph{word2vec} did
not significantly change the relative positions of words. This suggests
that $\model$ also captures the semantic relationship between words.

\begin{table*}
\begin{centering}
\begin{tabular}{|>{\raggedright}p{0.3\textwidth}|>{\raggedright}p{0.3\textwidth}|>{\raggedright}p{0.3\textwidth}|}
\hline 
\emph{Single delivery by cesarean section} & \emph{Type 2 diabetes mellitus}  & \emph{Atrial fibrillation and flutter }\tabularnewline
\hline 
\textbf{Diagnoses}: & \textbf{Diagnoses}: & \textbf{Diagnoses}:\tabularnewline
~~Maternal care for disproportion

~~Placenta praevia

~~Complications of puerperium

~~Failed induction of labor 

~~Diabetes mellitus in pregnancy  & ~~Personal history of medical treatment

~~Presence of cardiac/vascular implants

~~Personal history of certain other diseases

~~Unspecified diabetes mellitus 

~~Problems related to lifestyle & ~~Paroxysmal tachycardia

~~Unspecified kidney failure

~~Cardiomyopathy

~~Shock, not elsewhere classified 

~~Other conduction disorders \tabularnewline
\hline 
\textbf{Procedures:} & \textbf{Procedures:} & \textbf{Procedures:}\tabularnewline
~~Cesarean section

~~Medical or surgical induction of labour

~~Manipulation of fetal presentation

~~Other procedures associated with delivery

~~Forceps delivery  & ~~Cerebral anesthesia

~~Other digital subtraction angiography

~~Examination procedures on uterus

~~Medical or surgical induction of labour

~~Coronary angiography & ~~Insertion or removal procedures on aorta

~~Electrophysiological studies {[}EPS{]}

~~Other procedures on atrium

~~Coronary artery bypass - other graft

~~Coronary artery bypass - saphenous vein\tabularnewline
\hline 
\end{tabular}
\par\end{centering}
\caption{Retrieving top 5 similar diagnoses and procedures.\label{tab:Top-5-randinit}}
\end{table*}

\subsection{Filter Responses and Motifs}

While the semantics in the previous sub-section reveal the global
relative relation between diseases and procedures, they do not explain
local interactions (e.g., motifs). Here we compute the local filter
responses per sentence, and from there, a collection of strong and
frequent motifs is derived.

Table~\ref{fig:motif-sentence} shows some sentences with strong
responses for Filter 1 and 4 for both risk and no-risk class. It can
be seen that the sub-sequences Z85.1163.1910 and 1066.1067.I21 respond
strongly for the positive class and contribute to the classification
result. The first sub-sequence is about cancer history (Z85), biopsy
procedure (1163) and cerebral anesthesia (1910). The other sub-sequence
is about heart attack (I21) and kidney-related procedures (1066 and
1067). 

\begin{table}
\begin{centering}
\begin{tabular}{ll}
Filter ID & Response within a (sub) sentence \tabularnewline
\hline 
 & \tabularnewline
1 (readmit) & \includegraphics[width=0.35\columnwidth]{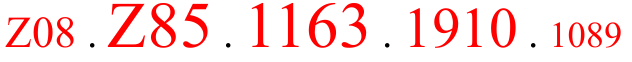}\tabularnewline
1 (no-risk) & \includegraphics[width=0.3\columnwidth]{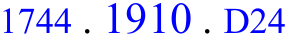}\tabularnewline
4 (readmit) & \includegraphics[width=0.7\columnwidth]{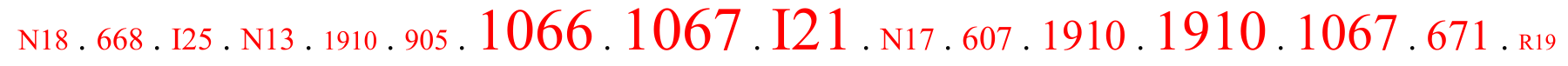}\tabularnewline
4 (no-risk) & \includegraphics[width=0.7\columnwidth]{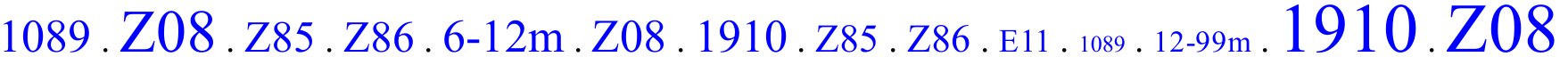}\tabularnewline
 & \tabularnewline
\end{tabular}
\par\end{centering}
\caption{Some sentences with strong responses for Filters 1 and 4. Code with
first letter is diagnosis, code with all numbers is procedure, code
ends with ``m'' is time-gap. The heights of the codes are proportional
to their response weights. The sub-sequence Z85.1163.1910 and 1066.1067.I21
response strongly to the positive class.\label{fig:motif-sentence}}
\end{table}

From strong and frequent filter responses in all sentences, we derive
the list of motifs. Table~\ref{tab:Retrieving-3-motifs} lists the
motifs with largest weights and highest frequency of occurrence for
code chapter E, I and O. The first motif of Filter 45 shows the pattern
that treatment removing toxic substances from the blood co-occurred
with care involving dialysis and readmission within 1 month. The second
motif in the same row discovers the pattern that type-I diabetes patients
involve in education about information and management of diabetes.
The third motif in the same row shows type-II diabetes patients readmit
within 1-3 months. Filter 26 demonstrates the co-occurrence of diseases
related to diabetes. The three motifs show that type-II diabetes patients
can have complications such as heart failure, vitamin D deficiency
and kidney failure. Filters 10 and 35 show diseases and treatments
related to the circulatory system, whereas pregnancy and birth related
motifs are shown in Filters 2 and 33 in the last two rows.

\begin{table*}
\begin{centering}
\begin{tabular}{|>{\centering}m{0.05\textwidth}|>{\raggedright}p{0.28\textwidth}|>{\raggedright}p{0.28\textwidth}|>{\raggedright}p{0.28\textwidth}|}
\hline 
\emph{Filter ID} & \multicolumn{3}{c|}{\emph{Motifs}}\tabularnewline
\hline 
45 & \textbf{0-1m 1060 Z49}

Time-gap

Haemoperfusion 

Care involving dialysis & \textbf{1916 E10 Z45}

Allied health intervention, diabetes education

Type 1 diabetes mellitus 

Adjustment and management of drug delivery or implanted device & \textbf{1-3m E11 Z45}

Time-gap

Type 2 diabetes mellitus

Adjustment and management of drug delivery or implanted device\tabularnewline
\hline 
26 & \textbf{E11 I48 I50}

Type 2 diabetes mellitus

Atrial fibrillation and flutter

Heart failure & \textbf{E11 E55 I48}

Type 2 diabetes mellitus

Vitamin D deficiency

Atrial fibrillation and flutter & \textbf{E11 I50 N17}

Type 2 diabetes mellitus

Heart failure

Acute kidney failure\tabularnewline
\hline 
10 & \textbf{1893 I48 K35}

Exchange transfusion

Atrial fibrillation and flutter 

Acute appendicitis & \textbf{1005 A41 I48}

Panendoscopy to ileum with administration of tattooing agent

Other sepsis 

Atrial fibrillation and flutter & \textbf{1-3m I48 Z45}

Time-gap 

Atrial fibrillation and flutter

Adjustment and management of drug delivery or implanted device\tabularnewline
\hline 
35 & \textbf{1909 727 I83}

Intravenous regional anesthesia

Interruption of sapheno-femoral and sapheno-popliteal junction varicose
veins

Varicose veins of lower extremities & \textbf{1620 I83 L57}

Excision of lesion(s) of skin and subcutaneous tissue of foot 

Varicose veins of lower extremities 

Skin changes due to chronic exposure to nonionising radiation & \textbf{1910 768 I83}

Sedation

Transcatheter embolisation of other blood vessels

Varicose veins of lower extremities\tabularnewline
\hline 
2 & \textbf{D68 O80 Z37}

Other coagulation defects

Single spontaneous delivery

Outcome of delivery & \textbf{1344 O75 O80}

Other suture of current obstetric laceration or rupture without perineal
involvement

Other complications of labor and delivery

Single spontaneous delivery & \textbf{1344 O75 O82}

Other suture of current obstetric laceration or rupture without perineal
involvement

Other complications of labour and delivery

Single delivery by caesarean section\tabularnewline
\hline 
33 & \textbf{1333 1340 O09}

Neuraxial block during labour and delivery procedure

Emergency lower segment caesarean section

Duration of pregnancy & \textbf{1340 O14 Z37}

Emergency lower segment caesarean section

Gestational {[}pregnancy-induced{]} hypertension with significant
proteinuria

Outcome of delivery & \textbf{1340 3-6m O34}

Emergency lower segment caesarean section

Time-gap

Maternal care for known or suspected abnormality of pelvic organs\tabularnewline
\hline 
\end{tabular}
\par\end{centering}
\caption{Retrieving 3 motifs for each of the 6 filters which have largest weights
and most frequent with code chapter O, I and E. \label{tab:Retrieving-3-motifs}}

\end{table*}

%% file: discuss.tex
We have presented $\model$, a new deep learning architecture that
provides an \emph{end-to-end} predictive analytics in healthcare services.
$\model$ reads directly from raw medical records and predicts future
outcomes. This departs from the traditional machine learning that
relies on expensive manual feature extraction. $\model$ learns to
extract meaningful features by itself without expert supervision.
This translates to uncovering the predictive local motifs in the space
of diseases and interventions. These capacities are not seen in existing
methods.

\paragraph{Significance}

$\model$ contributes to the growing literature of predictive medicine
in multiple ways. First, it is able to uncover the underlying space
of diseases and interventions, showing the relationships between them.
The largest disease cluster in Fig.~\ref{fig:disease-dist} suggests
that diseases may interact in a complex way, and current representation
of disease hierarchies such as those in ICD-10 may not reflect the
true nature of medical disorders. Second, $\model$ detects predictive
motifs of comorbidity, care patterns and disease progression. The
motifs suggest a new look into the complex interactions between diseases
and between the diseases and cares. Third, similar patients can be
retrieved not just using past history, but from likelihood of future
risks as well. This would, for example, help to quickly identify an
effective treatment regime based on similar patients who responded
well to the treatment, or to alert the care team of a potential risk
based on similar patients who had these before. Finally, $\model$
predicts the future risk for a patient and explains why (through means
of motifs responses), which is the core of modern prospective healthcare. 

With these capabilities, $\model$ can enable targeted monitoring,
treatments and care packaging. This is highly important for chronic
disease management that requires an on-going care and evaluation.
For health services, a high predictive accuracy of risk will lead
to better resources prioritizing and allocation. For patients, accurate
risk estimation is an important step toward personalized care. Patients
and family will be promoted to become more aware of the conditions
and risk, leading to proactive health management and help seeking.
$\model$ is generic and it can be implemented on existing EMR systems.
This will enable innovative healthcare practices for better efficiency
and outcomes to occur. For example, doctors, when seeing a patient,
may consult the machine for a second opinion, with a \emph{transparent,
evidence-based reasoning}. Because they do not miss any piece of information
in the database, they are less likely to overlook important signals.

\paragraph{Comparison to recent work on medical records}

Deep learning in healthcare has recently attracted great interest.
The most popular application is medical imaging using CNNs \cite{cirecsan2013mitosis},
motivated by the recent successes in cognitive vision \cite{he2015delving,krizhevsky2012imagenet,lecun2015deep}.
However, there has been limited work on non-cognitive modalities.
On time-series data (e.g., ICU measurements), the main difficulty
is the handling missing data with recent work of \cite{che2016recurrent,lasko2013computational,lipton2016directly,razavian2015temporal}.
In \cite{lasko2013computational}, time-series are modeled using autoencoders
(an unsupervised feedforward net) to discover meaningful phenotypes.
In \cite{che2016recurrent,lipton2016directly}, recurrent nets are
used, and in \cite{razavian2015temporal}, a convolutional net is
employed. $\model$ can be applied on these data, following a discretization
of continuous signals into discrete words (e.g., through cut-points).

On routine medical records, $\model$ is the only method that employs
convolutional nets but there exist alternative architectures. Feedforward
nets have been used \cite{liang2014deep,futoma2015comparison,miotto2016deep}.
Recurrent neural networks (RNN) on medical records include Doctor
AI \cite{choi2015doctor} and DeepCare \cite{pham2016deepcare}. Doctor
AI is a RNN adapted for medical events, where both next events and
time-gaps are predicted. DeepCare is a sophisticated model that represents
time-gaps using a parametric model. Similar to our observation, the
authors of DeepCare also noticed an interesting analogy between natural
languages and EMR, where EMR is similar to a sentence, and diagnoses
and interventions play the role of nouns and modifiers. While DeepCare
is powerful on long records, it is less effective in short records,
e.g., those with only one or two admissions. $\model$, on the other
hand, does not suffer from this limitation. Stochastic deep neural
nets such as deep Boltzmann machines are used in \cite{mehrabi2015temporal}.
Deep non-neural nets have also been suggested in \cite{henao2016electronic}.
These methods are likely to be expensive to train and produce prediction.

Embedding of medical concepts has been proposed in contemporary work
\cite{choi2016multi,choi2016learning,pham2016deepcare,tran2015learning}.
In \cite{choi2016learning}, medical concepts are embedded using \emph{word2vec}
\cite{mikolov2014word2vec}, ignoring time gaps. The \emph{Med2Vec}
in \cite{choi2016multi} extends \emph{word2vec} to embed visits.
Both \emph{word2vec} and \emph{Med2Vec} model local collocations,
but do not explicitly model motifs (with precise relative positions).
In \cite{tran2015learning}, a global model known as \emph{eNRBM}
embeds patients into vectors via regularized nonnegative restricted
Boltzmann machines \cite{tu_truyen_phung_venkatesh_acml13}. Local
motifs are not modeled and and variable record length and time gaps
are not properly handled. Discovering local motifs by means of convolutions
has been suggested in \cite{wang2012towards} through matrix factorization.
However, the work does not do prediction.

\paragraph{Limitations and future work}

There are rooms for future work. First, long-term dependencies are
simply captured through a max-pooling operation. This is rather simplistic
due to a complex dynamic between care processes and disease processes
\cite{pham2016deepcare}. A better model should pool information that
is time-sensitive (e.g., recent events are more important to distant
ones). At present, $\model$ works exclusively on recorded events
such as diagnoses and interventions. Integration with clinical narrative
would be highly useful because rich information is buried in unstructured
text. This can be done in the same framework of $\model$ because
of the sequential nature of text. Our evaluation has been limited
to a common risk known as unplanned readmission. However, $\model$
is not limited to any specific type of future risk. It can be well
applied to predicting the onset or progression of a disease.